# An Ontology-based Knowledge Management System for Industry Clusters

Pradorn Sureephong[1], Nopasit Chakpitak[1], Yacine Ouzrout[2], Abdelaziz Bouras[2]

[1]Department of Knowledge Management, College of Arts, Media and Technology, Chiang Mai University, Chiang Mai, Thailand. {dorn | nopasit}@camt.info
[2]LIESP, University Lumiere Lyon 2, Lyon, France, {yacine.ouzrout | abdelaziz.bouras}@univ-lyon2.fr

**Abstract**

Knowledge-based economy forces companies in every country to group together as a cluster in order to maintain their competitiveness in the world market. The cluster development relies on two key success factors which are knowledge sharing and collaboration between the actors in the cluster. Thus, our study tries to propose a knowledge management system to support knowledge management activities within the cluster. To achieve the objectives of the study, ontology takes a very important role in the knowledge management process in various ways; such as building reusable and faster knowledge-bases and better ways of representing the knowledge explicitly. However, creating and representing ontology creates difficulties to organization due to the ambiguity and unstructured nature of the source of knowledge. Therefore, the objectives of this paper are to propose the methodology to capture, create and represent ontology for organization development by using the knowledge engineering approach. The handicraft cluster in Thailand is used as a case study to illustrate our proposed methodology.

*Keywords: Ontology, Semantic, Knowledge Management System, Industry Cluster*

## 1.1 Introduction

In the past, the three production factors (Land, Labor and Capital) were abundant, accessible and were considered as the reason of economic advantage, knowledge did not get much attention [1]. Nowadays, it is the knowledge-based economy era which is affected by the increasing use of information technologies. Thus, previous production factors are currently no longer enough to sustain a firm's competitive advantage; knowledge is being called on to play a key role [2]. Most industries try to use available information to gain more competitive advantages than others. Knowledge-based economy is based on the production, distribution and use of knowledge and information [3]. The study of Yoong and Molina [1] assumed that one way of surviving in today's turbulent business environment for business organizations is to form strategic alliances or mergers with other similar or complementary business companies. The conclusion of Yoong and Molina's study supports the idea of industry cluster [3] which is proposed by Porter in 1990.

The objectives of the grouping of firms as a cluster are maintaining the collaboration and sharing of knowledge among the partners in order to gain competitiveness in their market. Therefore, Knowledge Management (KM) becomes a critical activity in achieving the goals. In order to manage the knowledge, ontology plays an important role in enabling the processing and sharing of knowledge between experts and knowledge users. Besides, it also provides a shared and common understanding of a domain that can be communicated across people and application systems. On the other hand, creating ontology for an industry cluster can create difficulties to the Knowledge Engineer (KE) as well, because of the complexity of the structure and time consumed. In this paper, we will propose the methodology for ontology creation by using knowledge engineering methodology in the industry cluster context.

## 1.2 Literature Review

### 1.2.1 Industry Cluster and Knowledge Mangement

The concept of the industry cluster was popularized by Prof. Michael E. Porter in his book "Competitive Advantages of Nations" [3] in 1990. Then, industry cluster becomes the current trend in economic development planning. However, there is considerable debate regarding the definition of the industry cluster. Based on Porter's definition of industry cluster [4], the cluster can be seen as a "*geographically proximate group of companies and associated institutions (for example universities, government agencies, and related associations) in a particular field, linked by commonalities and complementarities*". The general view of industry cluster map is shown in figure 1.1. Until now, literature of the industry cluster and cluster building has been rapidly growing both in academic and policy-making circles [5].

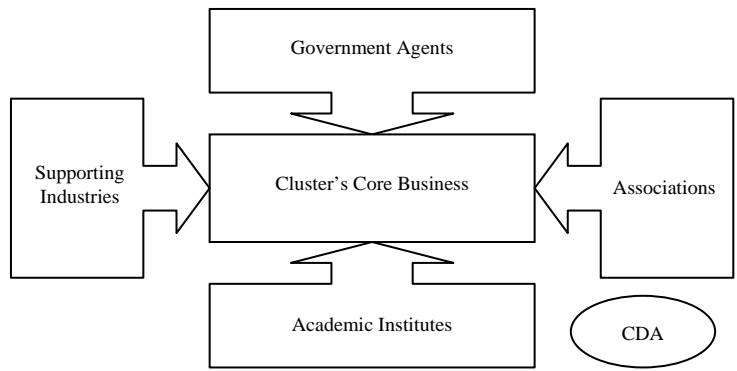

**Figure 1.1.** Inustry Cluster Map

After the concept of industry cluster [3] was tangibly applied in many countries, companies in the same industry tended to link to each other to maintain their competitiveness in their market and to gain benefits from being a member of the cluster. From the study of ECOTEC in 2005[6] regarding the critical success factors in cluster development, the two critical success factors are *collaboration* in networking partnership and *knowledge creation* for innovative technology in the cluster which are about 78% and 74% of articles mentioned as success criteria accordingly. This knowledge is created through various forms of local inter-organizational collaborative interaction [7]. They are collected in the form of tacit and explicit knowledge in experts and institutions within cluster. We applied knowledge engineering techniques to the industry cluster in order to capture and represent the tacit knowledge in the explicit form.

**1.2.2 Knowledge Engineering Techniques**

Initially knowledge engineering was just a field of the artificial intelligence. It was used to develop knowledge-based systems. Until now, knowledge engineers have developed their principles to improve the process of knowledge acquisition since last decade [8]. These principles are used to apply knowledge engineering in many actual environment issues. Firstly, there are different types of knowledge. This was defined as "know what" and "know how" [9] or "explicit" and "tacit" knowledge from Nonaka's definition [10] Secondly, there are different type of experts and expertise. Thirdly, there are many ways to present knowledge and use of knowledge. Finally, the use of structured method to relate the difference together to perform knowledge oriented activity [11].

In our study, many knowledge engineering methods have been compared [12] in order to select a suitable method to be applied to solve the problem of industry cluster development; i.e. SPEDE, MOKA, CommonKADS. We adopted CommonKADS methodology because it provides sufficient tools; such as a model suite (figure 1.2) and templates for different knowledge intensive tasks.

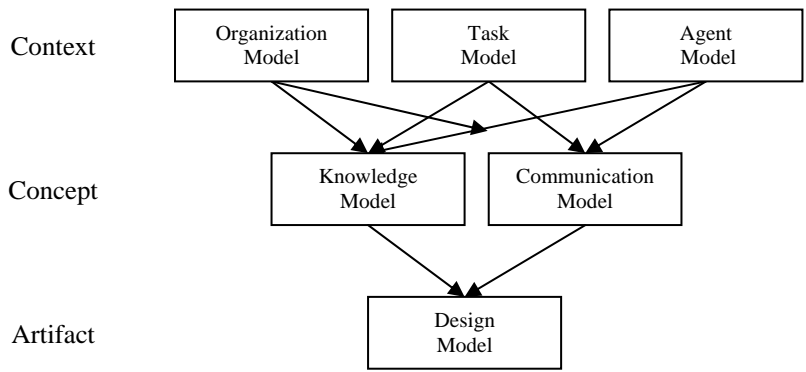

**Figure 1.2.** CommonKADS Model Suite

**1.2.2 Ontology and Knowledge Management**

The definition of ontology by Gruber (1993) [13] is *"explicit specifications of a shared conceptualization"*. A *conceptualization* is an abstract model of facts in the world by identifying the relevant concepts of the phenomenon. *Explicit* means that the type of concepts used and the constraints on their use are explicitly defined. *Shared* reflects the notion that an ontology captures consensual knowledge, that is, it is not private to the individual, but accepted by the group.

Basically, the role of ontology in the knowledge management process is to facilitate the construction of a domain model. It provides a vocabulary of terms and relations in a specific domain. In building a knowledge management system, we need two types of knowledge [14]:

*Domain knowledge:* Knowledge about the objective realities in the domain of interest (Objects, relations, events, states, causal relations, etc. that are obtained in some domains)

*Problem-solving knowledge:* Knowledge about how to use the domain knowledge to achieve various goals. This knowledge is often in the form of a problem-solving method (PSM) that can help achieve the goals in a different domain.

In this study, we focus on ontology creation and representation by adopting knowledge engineering methodology to support both dimensions of knowledge. We use the ontology as a main mechanism to represent information and knowledge, and to define the meaning of terms used in the content language and the relation in the knowledge management system.

## 1.3 Methodology

Our proposed methodology divides ontology into three types: generic ontology, domain ontology and task ontology. *Generic ontology* is the ontology which is re-useable across the domain, e.g. organization, product specification, contact, etc. *Domain ontology* is the ontology defined for conceptualizing on the particular domain, e.g. handicraft business, logistic, import/export, marketing, etc. *Task ontology* is the ontology that specifies terminology associated with the type of tasks and describes the problem solving structure of all the existing tasks, e.g. paper production, product shipping, product selection, etc.

In our approach to implement ontology-based knowledge management, we integrated existing knowledge engineering methodologies and ontology development processes. We adopted CommonKADS for knowledge engineering methodology and OnToKnowledge (OTK) methodology for ontology development. Figure 1.3 shows the assimilation of CommonKADS and On-To-Knowledge (OTK) [15].

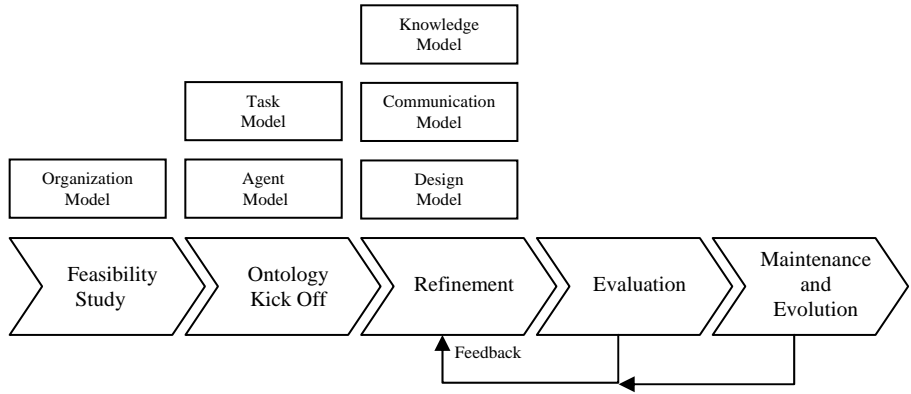

**Figure 1.3.** Steps of OTK methodology and CommonKADS model suite

### 3.1 Feasibility Study Phase

The feasibility study serves as decision support for an economical, technical and project feasibility study, in order to select the most promising focus area and target solution. This phase identifies problems, opportunities and potential solutions for the organization and environment. Most of the knowledge engineering methodologies provide the analysis method to analyze the organization before the knowledge engineering process. This helps the knowledge engineer to understand the environment of the organization. CommonKADS also provides context levels in the model suite (figure 1.2) in order to analyze organizational environment and the corresponding critical success factors for a knowledge system [16]. The organization model provides five worksheets for analyzing feasibility in the organization as shown in figure 1.4.

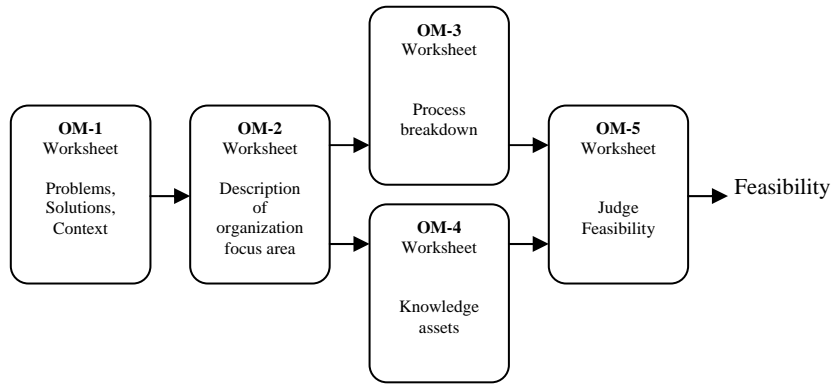

**Figure 1.4.** Organization Model Worksheets

The Knowledge engineer can utilize OM-1 to OM-5 worksheets for interviewing with knowledge decision makers of organizations. Then, the outputs from OM are a list of knowledge intensive tasks and agents which are related to each task. Then, KE could interview experts in each task using TM and AM worksheets for the next step. Finally, KE validates the result of each module with knowledge decision makers again to assess impact and changes with the OTA worksheet.

### 1.3.2 Ontology Kick Off Phase

The objective of this phase is to model the requirements specification for the knowledge management system in the organization. The Ontology Requirement Support Document (ORSD) [17]guides knowledge engineers in deciding about inclusion and exclusion of concepts/relations and the hierarchical structure of the ontology. It contains useful information, i.e. Domain and goal of the ontology, Design guidelines, Knowledge source, User and usage scenario, Competency questions, and Application support by the ontology[15].

Task and Agent Model are separated in to TM-1, TM-2 and AM worksheets. They facilitate KE to complete the ORSD. The TM-1 worksheet identifies the features of relevant tasks and knowledge sources available. TM-2 worksheet concentrates in detail on bottleneck and improvement relating to specific areas of knowledge. AM worksheet lists all relevant agents who possess knowledge items such as domain experts or knowledge workers.

### 1.3.3 Refinement Phase

The goal of the refinement phase is to produce a mature and application-oriented target ontology according to the specification given by the kick off phase [18]. The main tasks in this phase are knowledge elicitation and formalization.

*Knowledge elicitation* process with the domain expert based on the initial input from the kick off phase is performed. CommonKADS provides a set of knowledge templates [11] in order to support KE to capture knowledge in different types of tasks. CommonKADS classify knowledge intensive tasks in two categories; i.e. analytic tasks and synthetic tasks. The first is a task regarding systems that pre-exist. In opposition, the synthetic task is about the system that does not yet exist. Thus, KE should realize about the type of task that he is dealing with. Figure 1.5 shows the different knowledge task types.

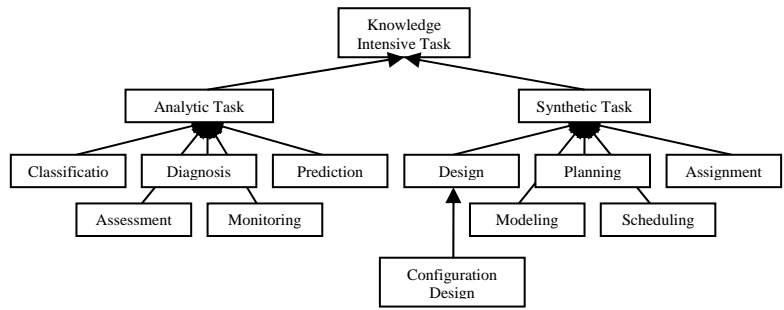

**Figure 1.5.** Knowledge-intensive task types based on the type of problem

*Knowledge formalization* is transformation of knowledge into formal representation languages such as Ontology Inference Layer (OIL) [19], depends on application. Therefore, the knowledge engineer has to consider the advantages and limitations of the different languages to select the appropriate one.

### 1.3.4 Evaluation Phase

The main objectives of this phase are to check, whether the target ontology suffices the ontology requirements and whether the ontology based knowledge management system supports or answers the competency questions, analyzed in the feasibility and kick off phase of the project. Thus, the ontology should be tested in the target application environment. A prototype should already show core functionalities of the target system. Feedbacks from users of the prototype are valuable input for further refinement of the ontology. [18]

### 1.3.5 Maintenance and Evolution Phase

The maintenance and evolution of an ontology-based application is primarily an organizational process [18]. The knowledge engineers have to update and maintain the knowledge and ontology in their responsibility. In order to maintain the knowledge management system, an ontology editor module is developed to help knowledge engineers.

## 1.4 Case Study

The initial investigations have been done with 10 firms within the two biggest handicraft associations in Thailand and Northern Thailand. *No*rthern *H*andicraft *M*anufacturer and *EX*porter (NOHMEX) association is the biggest handicraft association in Thailand which includes 161 manufacturers and exporters. Another association which is the biggest handicraft association in Chiang Mai is named Chiang Mai Brand which includes 99 enterprises. It is a group of qualified manufacturers who have capability to export their products and pass the standard of Thailand's ministry of commerce.

The objective of this study is to create a Knowledge Management System (KMS) for supporting this handicraft cluster. One of the critical tasks to implement this system is creating ontologies of the knowledge tasks. Because, ontology is recognized as an appropriate methodology to accomplish a common consensus of communication, as well as to support a diversity of activities of KM, such as knowledge repository, retrieval, sharing, and dissemination [20]. In this case, knowledge engineering methodology was applied for ontology creation in the domain of Thailand's handicraft cluster.

**Domain Ontology:** can be created by using three models in context level of model suite; i.e. organization model, task model and agent model. At the beginning of domain ontology creation, we adopt generic ontology plus acquired information from the worksheets as an outline. Then, the more information that can be acquired from organization and environment, the more domain-oriented ontology can be filled-in.

**Task Ontology:** specifies terminology associated with the type of tasks and describes the problem solving structure. The objective of knowledge engineering methods is to solve problems in a specific domain. Thus, most of knowledge engineering approaches provide a collection of predefined sets of model elements for KE [16]. CommonKADS methodology also provides a set of templates in order to support KE to capture knowledge in different types of tasks. As shown in figure 1.5, there are various types of knowledge tasks that need different ontology. Thus, KE has to select the appropriate template in order to capture right knowledge and ontology. For illustration, we will use *classification template* for analytic task as an example for task ontology creation. Figure 1.6 shows the inferences structure for classification method (left side) and task ontology (right side).

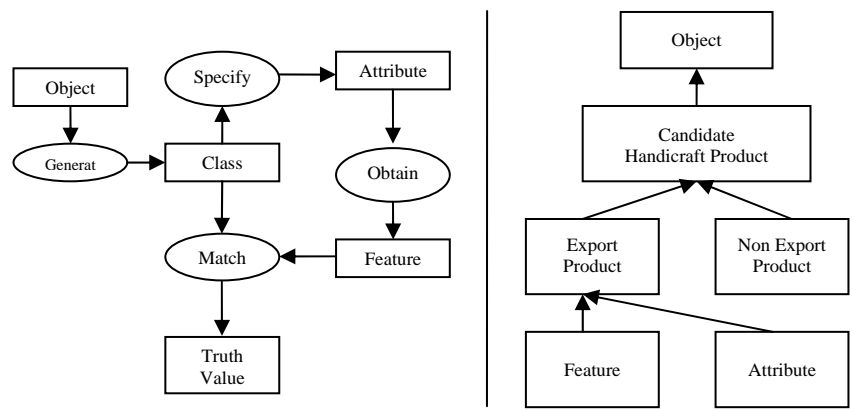

**Figure 1.6.** CommonKADS classification template and task ontology

In the case study of a handicraft cluster, one of the knowledge intensive tasks is about product selection for exporting. Not all handicraft products are exportable due to their specifications, function, attributes, etc. Moreover, there are many criteria to select a product to be exported to specific countries. So we defined the task ontology of the product selection task (see the right side of figure 1.6).

## 1.5 Conclusion

The most important role of ontology in knowledge management is to enable and to enhance knowledge sharing and reusing. Moreover, it provides a common mode of communication among the agents and knowledge engineer [14]. However, the difficulties of ontology creation are claimed in most literature. Thus, this study focuses on creating ontology by adopting the knowledge engineering methodology which provides tools to support us for structuring knowledge. Thus, ontology was applied to help Knowledge Management System (KMS) for the industry cluster to achieve their goals. The architecture of this system  consists of three parts, knowledge system, ontology, and knowledge engineering. Hence, the proposed methodology was used to create ontology in the handicraft cluster context. During the manipulation stage, when users accesses the knowledge base, the ontology can support tasks of KM as well as searching. The knowledge base and the ontology is linked one to another via the ontology module. In the maintenance stage, knowledge engineers or domain experts can add, update, revise, and delete the knowledge or domain ontology via knowledge acquisition module [21].

To test and validate our approach and architecture, we used the handicraft cluster in Thailand as a case study. In our perspectives of this study, we will finalize the specification of the shareable knowledge/information and the conditions of sharing among the cluster members. Then, we will capture and maintain the knowledge (for reusing knowledge when required) and work on the specific infrastructure to enhance the collaboration. At the end of the study, we will develop the knowledge management system for the handicraft cluster relating to acquiring requirements specification from the cluster.

## 1.6 References

[1] Young P, Molina M, (2003) Knowledge Sharing and Business Clusters, In: 7th Pacific Asia Conference on Information Systems, pp.1224-1233.
[2] Romer P, (1986) Increasing Return and Long-run Growth, Journal of Political Economy, vol. 94, no.5, pp.1002-1037.
[3] Porter M E, (1990) Competitive Advantage of Nations, New York: Free Press.
[4] Porter M E, (1998) On Competition, Boston: Harvard Business School Press.
[5] Malmberg A, Power D, (2004) (How) do (firms in) cluster create knowledge?, in DRUID Summer Conference 2003 on creating, sharing and transferring knowledge, Copenhagen, June 12-14.
[6] DTI, (2005) A Practical Guide to Cluster Development, Report to Department of Trade and Industry and the English RDAs by Ecotec Research & Consulting.
[7] Malmberg A, Power D, On the role of global demand in local innovation processes: Rethinking Regional Innovation and Change, Shapiro P, and Fushs G, Dordrecht, Kluwer Academic Publishers.
[8] Chua A, (2004) Knowledge management system architecture: a bridge between KM consultants and technologist, International Journal of Information Management, vol. 24, pp. 87-98.
[9] Lodbecke C, Van Fenema P, Powell P, Co-opetition and Knowledge Transfer, The DATA BASE for Advances in Information System, vol.30, no. 2, pp.14-25.
[10] Nonaka I, Takeuchi H, (1995) The Knowledge-Creating Company, Oxford University Press, New York.
[11] Shadbolt N, Milton N, (1999) From knowledge engineering to knowledge management, British Journal of Manage1ment, vol. 10, no. 4, pp. 309-322, Dec.
[12] Sureephong P, Chakpitak N, Ouzrout Y, Neubert G, Bouras A, (2006) Economic based Knowledge Management System for SMEs Cluster: case study of handicraft cluster in Thailand. SKIMA Int. Conference, pp.10-15.
[13] Gruber TR, (1991) The Role of Common Ontology in Achieving Sharable, Reusable Knowledge Bases, In J. A. Allen, R. Fikes, & E. Sandewall (Eds.), *Principles of Knowledge Representation and Reasoning: Proceedings of the Second International Conference,* Cambridge, MA, pp. 601-602.

[14] Chandrasekaran B, Josephson, JR, Richard BV, (1998) Ontology of Tasks and Methods, In Workshop on Knowledge Acquisition, Modeling and Management (KAW'98), Canada.
[15] Sure Y, Studer R, (2001) On-To-Knowledge Methodology, evaluated and employed version. On-To-Knowledge deliverable D-16, Institute AIFB, University of Karlsruhe.
[16] Schreiber A Th, Akkermans H, Anjewerden A, de Hoog R, Shadbolt N, van de Velde W, Wielinga B, (1999) Knowledge Engineering and Management: The CommonKADS Methodology, The MIT Press.
[17] Sure Y, Studer R, (2001) On-To-Knowledge Methodology, final version. On-To-Knowledge deliverable D-18, Institute AIFB, University of Karlsruhe.
[18] Staab S, Schnurr HP, Studer R, Sure Y, (2001) Knowledge processes and ontologies, IEEE Intelligent Systems, 16(1):26-35.
[19] Fensel, Harmelen Horrocks (OIL)
[20] Gruber T R, (1997) Toward principles for the design of ontologies used for knowledge sharing, Int. J Hum Comput Stud, vol. 43, no. 5-6, pp.907-28.
[21] Chau K W, (2007) An ontology-based knowledge management system for flow and water quality modeling, Advance in Engineering Software, vol. 38, pp. 172-181.